# Detection of Problem Gambling with Less Features Using Machine Learning Methods

Yang Jiao[1], Gloria Wong-Padoongpatt[2], Mei Yang[1]
[1]Department of Electrical and Computer Engineering
[2] Department of Psychology
University of Nevada, Las Vegas

**Abstract**

Analytic features in gambling study are performed based on the amount of data monitoring on user daily actions. While performing the detection of problem gambling, existing datasets provide relatively rich analytic features for building machine learning based model. However, considering the complexity and cost of collecting the analytic features in real applications, conducting precise detection with less features will tremendously reduce the cost of data collection. In this study, we propose a deep neural networks *PGN4* that performs well when using limited analytic features. Through the experiment on two datasets, we discover that PGN4 only experiences a mere performance drop when cutting 102 features to 5 features. Besides, we find the commonality within the top 5 features from two datasets.

## 1. Introduction

While the Internet gambling (also called online gambling) has grown dramatically during the past two decades, the issue of problem gambling has attracted massive attention from the community of gambling research because of the significant negative impact it causes from the perspectives of individual and public health (Deng et al., 2018). To detect problem gambling, online gambling behaviors which inherently link to individual accounts are monitored and recorded over time (Griffiths, 2012). These behavioral datasets are transformed into analytic datasets and features which conclude the user behaviors into information such as betting amount, betting frequency, frequent games, account actions, etc.

Table 1 Machine learning approaches for addiction research

| Category | Method | Description | Ref. |
|---|---|---|---|
| Supervised learning | Regression | Regression models which include logistic regression, and multiple types of penalized regression (Regis, Lasso, Elastic Net), optimize several parameters when training | Acion et al. (2017) Soussia and Rekik (2018) Rish et al. (2016) |
| | Support Vector Machine (SVM) | SVM is a discriminative classifier that determines the separating hyperplane between data classes. While training, SVM maximizes the distance between data and the hyperplane to optimize the classification. | Soussia and Rekik (2018) Rish et al. (2016) |
| | Trees | Demonstrating an advantage on visualizing a decision making process, decision trees build tree-like graphs to separate data. CHAID analyzes the relation between features in decision tree. | Braverman et al. (2013) Rish et al. (2016) Rho et al. (2016) |
| | Random Forests (RF) | Considering a single tree may not be sufficient, random forests implement multiple decision trees to perform classification. | Soussia and Rekik (2018) Rish et al. (2016) |
| | Naive Bayes | Naïve Bayes is a generative model that assumes all features are independent. | Rish et al. (2016) |
| | Boosting | Based on a similar idea as RF, boosting methods compose multiple types of classifier to improve the performance. | - |
| | Discriminant analysis | Discriminant analysis finds a linear combination of features that separates two or more classes. | Gray et al. (2012) Rish et al. (2016) |
| | Neural Networks | Neural networks are a set of algorithms that implement layers of neurons to contain weights and achieve non-linear transformation. | Acion et al. (2017) Soussia and Rekik (2018) |
| | Deep Neural Networks | Deep Neural Networks stack the convolutional layers to distill high-level and abstract features. | - |
| Unsupervised learning | K-means | K-means is a non-parameterized algorithm that automatically clusters data into N groups. | Braverman & Shaffer (2012) Gray et al. (2015) |
| Reinforced learning | Q-learning | Q-learning is a reinforcement learning algorithm that seeks to find the best action to take given the current state without a policy. | Baker et al. (2020) |

However, analytic features require massive user data monitoring and therefore costly to obtain. In real applications, we also find that the available analytic features vary tremendously between datasets (Gray et al., 2012, Braverman et al., 2013, Braverman & Shaffer, 2012). To accommodate small datasets and reduce the cost of feature obtaining, we propose to study problem gambling detection with less or limited features using machine learning approaches.

In the last decades, machine learning methods have dominated the dataset analysis for addiction research, including problem gambling or high-risk gambler detection (Mak et al., 2019). Supervised, unsupervised, and reinforcement learning are three categories of machine learning approaches. As the most commonly applied machine learning type, supervised learning employs raw data and annotated ground truth to train classifiers (or regressors). It shows good promise on the quality and speed of convergence. Unsupervised learning avoids the labor cost of annotation and draw inferences from datasets consisting of input data without labeled responses. Reinforcement learning aims to optimize an agent to take action to respond to the current state. Table 1 presents a technique taxonomy for machine learning approaches on addiction research.

If we consider the raw analytic features as low-level features, one noticeable drawback of the aforementioned machine learning approaches is that they focus on low-level or the combination of low-level features and ignore the possibility of continuous combining low-level features into abstract (high-level) features. As a subfield of machine learning, deep neural networks classifiers (LeCun et al., 2015) harness multi-layered neural networks to automatically convert data into abstract representations via adjusting their weights. Other than demonstrating power in language and image processing, one dimension deep neural networks as 1-D CNN has been widely applied to time series data analysis (Kiranyaz et al., 2019), for example, signal analysis. However, 1-D CNN has rarely been applied to analytic data.

To obtain a rich feature space, 1-D CNN can combine low-level features from data within a local time frame into abstract features. The enriched abstract feature space will benefit the application with limited features. Therefore, referring to the concept of deep neural networks, the hypothesis is that 1-D CNN will extract abstract features from raw analytic features and will boost the performance of problem gambling detection with limited features.

There are two aspects of performance boosting: (1) the overall performance with full features, and (2) the overall performance with limited features. Considering the cost and complexity of collecting rich analytic features in real applications, the focus of this work is to study the approach that will boost the overall performance with limited features.

2. Method

Deep neural network classifiers (Jiao et al., 2018, 2019) are implemented by stacking varying types of layers by restrictive rules. Major layers in deep neural network classifier are listed below.

- Convolutional layer: As one of the major components in deep neural network classifier, convolutional layer distills abstract features with multiple sliding filters with weights which are optimized during training. To keep the output size, zeroes are commonly padded around the sample.

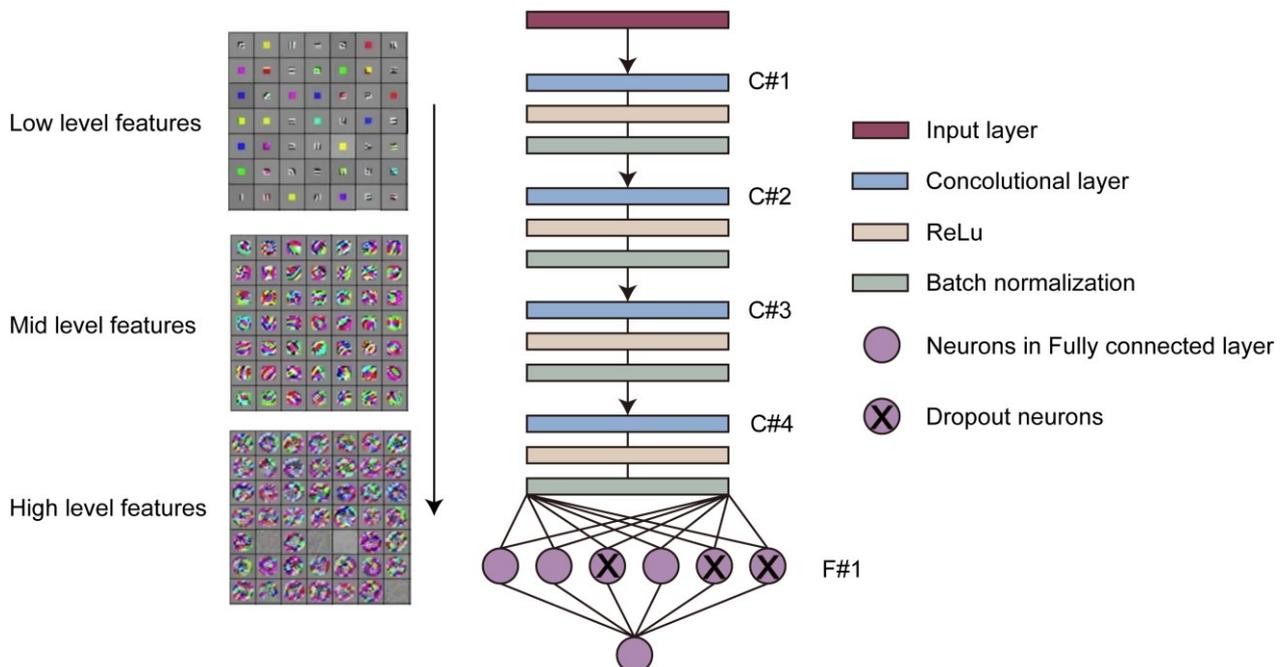

Figure 1 PGN4 architecture

Table 2 Parameters of layers in PGN4

| Layers | Parameters |
|---|---|
| 1D Convolutional layer C#1 | filter size=3; filter channel=16; stride=1; padding=same |
| 1D Convolutional layer C#2 | filter size=3; filter channel=16; stride=2; padding=same |
| 1D Convolutional layer C#3 | filter size=3; filter channel=32; stride=1; padding=same |
| 1D Convolutional layer C#4 | filter size=3; filter channel=32; stride=2; padding=same |
| Fully connected layerF #1 | 128 neurons |

- Pooling layer: Pooling layer operates on each feature map independently and aggressively reduces the spatial size of the abstract features, which consequently reduces the computational cost. Max pooling is the most popular pooling function.
- Fully connected layer: To transfer abstract feature maps into a classification score, a fully connected layer flattens the feature maps into a vector of neurons to perform a nonlinear transformation.
- Activation layer: To avoid gradient exploding, activation layer overlaps a nonlinear transformation on the output of the previous layer to map the output into a restricted range. Some common functions are ReLu, and Leak ReLu.
- Batch normalization layer: This layer aims to accelerate the training and against overfitting by normalizing a layer input into a restricted range.
- Dropout layer: Dropout layer randomly mutes some neurons to force robust learning.

Typically, several (one to three) convolutional layers are connected sequentially to perform abstract feature extraction. An activation layer and a batch normalization layer follow a convolutional layer to constrain the convolutional output. A pooling layer performs to reduce the output size.

The depth of the 1-D CNN depends on the raw feature vector size and the data size. Deeper networks have more weight parameters to train, and consequently, require exponentially increased data. Considering the existing behavioral datasets for online gambling, we implement a four-convolutional-layer CNN, namely *Problem Gambling Net 4* or *PGN4*. In PGN4, the convolutional layers with stride size 2 replace pooling layers to reduce the feature map spatial size. Figure 1 shows the PGN4 architecture and the progressive distilling of abstract features. The parameter design for layers in PGN4 is shown in Table 2.

### 2.1 Feature selection

To boost the detection performance with limited features, a feature selection is conducted based on feature correlation analysis. Through the selection, we evaluate the PGN4 with 5, 10, 20, 50, and full features. Because PGN4 distills abstract features by sliding filters, the arrangement of the raw feature vector will make a difference in abstract features and ultimately impact the detection performance. Via Algorithm 1, we arrange the most correlated features adjacently.

---

**Algorithm 1**

**Input:** Behavioral features vector $f$, Number of feature selections $N$, problem gambler flags $FL$
**Output:** Rearranged behavioral features $f'$
1: for $f_i$ in $f$
2:   Compute the correlation between features and flags
     $C(i) = \text{corrcoef}(f_i, FL)$
3:   for $f_j$ in $f$

---

Table 3 Reason of the RG Program flags

| Reason | Proportion |
|---|---|
| Account closure/reopening due to problem gambling. | 40%-45% |
| The user reports a problem. | 14%-16% |
| The user requests a limit change. | 15%-22% |
| The user requests to block one or multiple but not all games due to problem gambling | 13%-15% |
| The user requests a higher personal deposit limit. | 4%-5% |
| The user heavily complains about fair play. | 2% |
| A third party contacts RG program to block a user account. | 0%-1% |
| The user cancels an out-payment after requesting it. | 0%-1% |
| The user requests to block an in-payment method. | 0%-1% |
| The user is under age. | 0%-1% |
| Others or unclassified | 0%-1% |

4:     Compute the correlation matrix between features
       $C_f(i,j) = \text{corrcoef}(f_i, f_j)$
5: end
6: end
7: sort all C, $C_f(i)$ in descending order as $C'$, $C_f'$
8: //The bow of candidate features are the top
   //correlated features with flags $f_b = C'(1:N)$
9: for $n = 1:N, m = 1:N$
10:   if $f_b(m)$ not in $f'$
11:     $f'(n) = f_b(m)$
    //We assign a most correlated feature adjacent
    if $C_f'(f'(n), 1)$ not in $f'$
12:     $f'(n+1) = C_f'(f'(n), 1)$
13:   end
14: end
15: until rearranged all candidate features in $f_b$ into $f'$

## 2.2 Training

The PGN4 is trained with, Adam optimizer and learning rate $2\times10^{-4}$ in 20 epochs. Adam optimizer

Table 4 Performance comparison on two datasets. Best performances are bold. Acc: Accuracy

| Feature | Approach | Dataset A | | | | Dataset B | | | |
|---|---|---|---|---|---|---|---|---|---|
| | | Acc | F1 Score | ROC AUC | PR AUC | Acc | F1 Score | ROC AUC | PR AUC |
| Full | PGN4 | 70.5% | 64.3% | 74.6% | 76.7% | **80.8%** | **82.3%** | **90.2%** | 89.5% |
| | SVM | 61.3% | 70.6% | 63.2% | 53.4% | 66.0% | 74.5% | 74.2% | 53.5% |
| | DT | 61.8% | 62.0% | 61.1% | 71.7% | 72.2% | 72.6% | 71.5% | 79.6% |
| | RF | 66.0% | 63.3% | 71.9% | 71.8% | 77.5% | 76.3% | 84.5% | 85.5% |
| | Ada | 70.2% | 67.9% | **77.1%** | **78.5%** | 78.8% | 77.9% | 85.2% | **87.3%** |
| | NN | **72.3%** | **71.9%** | 74.1% | 77.9% | 68.0% | 75.2% | 88.8% | 89.5% |
| 50 | PGN4 | 69.8% | 62.1% | 74.7% | 76.5% | - | - | - | - |
| | SVM | 60.0% | **70.3%** | 59.3% | 49.8% | - | - | - | - |
| | DT | 60.7% | 62.1% | 60.7% | 71.5% | - | - | - | - |
| | RF | 67.2% | 62.5% | 72.1% | 72.3% | - | - | - | - |
| | Ada | 69.4% | 68.6% | **76.6%** | **78.3%** | - | - | - | - |
| | NN | **72.4%** | 70.0% | 76.3% | 77.8% | - | - | - | - |
| 20 | PGN4 | **69.0%** | 62.4% | 73.2% | 75.6% | 80.3% | 77.8% | **90.2%** | 90.1% |
| | SVM | 60.8% | **70.1%** | 63.2% | 54.1% | 64.8% | 73.9% | 74.2% | 53.2% |
| | DT | 60.7% | 60.5% | 60.3% | 70.6% | 71.9% | 72.5% | 71.1% | 79.3% |
| | RF | 66.9% | 61.4% | 69.3% | 70.1% | 79.1% | 78.0% | 84.8% | 85.7% |
| | Ada | 68.1% | 65.7% | **74.1%** | 76.9% | 78.7% | 78.0% | 85.6% | 87.8% |
| | NN | 68.1% | 67.9% | 71.7% | 75.4% | **81.7%** | **82.5%** | 89.4% | 90.0% |
| 10 | PGN4 | **67.9%** | 65.7% | 73.9% | 74.8% | **80.5%** | 78.6% | **88.0%** | **88.0%** |
| | SVM | 66.7% | **66.7%** | 69.3% | 64.6% | 65.7% | 72.7% | 69.2% | 56.9% |
| | DT | 62.1% | 58.7% | 59.6% | 69.2% | 69.3% | 69.9% | 69.0% | 77.8% |
| | RF | 66.7% | 63.5% | 64.9% | 66.2% | 75.0% | 73.6% | 79.3% | 81.4% |
| | Ada | 67.1% | 66.3% | 71.6% | 72.4% | 75.1% | 73.1% | 81.5% | 84.9% |
| | NN | 66.8% | 65.5% | **74.1%** | **75.0%** | 75.6% | **79.5%** | 87.2% | 86.6% |
| 5 | PGN4 | **68.8%** | **67.8%** | **74.1%** | **75.0%** | **79.2%** | **79.6%** | **87.9%** | **87.8%** |
| | SVM | 67.5% | 65.9% | 68.0% | 66.5% | 74.2% | 73.7% | 75.6% | 73.8% |
| | DT | 61.3% | 55.8% | 58.0% | 68.0% | 66.8% | 65.4% | 64.5% | 74.1% |
| | RF | 66.2% | 62.6% | 64.6% | 66.4% | 74.2% | 72.0% | 76.6% | 79.1% |
| | Ada | 66.3% | 65.7% | 69.6% | 72.0% | 74.9% | 73.6% | 80.6% | 84.2% |
| | NN | 67.4% | 67.3% | 74.0% | 74.4% | 75.0% | 79.1% | 86.4% | 86.1% |

(Kingma & Ba, 2014) is a state-of-the-art model optimizer which calculates an exponential moving average of the gradient and the squared gradient from the training loss of a minibatch of samples, and the parameters beta1 and beta2 control the decay rates of these moving averages. The loss function used to compute the training loss is binary cross-entropy.

## 3. Performance evaluation

In this works, we collect two public datasets to evaluate the performance of the proposed PGN4. Both datasets include multiple modalities of online gambling such as live action sports gambling, fix-odds sporting betting, casino, poker, and games like backgammon. Excluding date and categorical features, Dataset A (Braverman et al., 2013) contains 102 numerical behavioral features of 4,056 users, and Dataset B (Gray et al., 2012) has 27 numerical behavioral features of 4,132 users, as 25% of data are randomly select for validation. In both datasets, the user behavioral data associate with the Internet betting service provider *bwin.party,* and the flags of problem gamblers are provided by the *Responsible Gambling* (RG) program. The RG program flags a user based on multiple reasons, as shown in Table 3.

### 3.1 Evaluation metrics

To comprehensively evaluate the performance of PGN4, we apply 4 evaluation metrics including accuracy, F1 score, Precision-Recall (PR) curve, and Receiver operating characteristic (ROC) curve because they perform fair evaluation for either balanced or imbalanced data considering both positives and negatives. The area-under-curve (AUC) represents the overall performance of a classifier in the PR curve and ROC curve evaluation.

### 3.2 Performance comparison

Table 4 lists the performance metrics of PGN4 and five methods in comparison. According to the result, when performing full features on problem gambling detection, PGN4 is not always the best classifier because the feature space is abundant with full analytic features. As such the rich abstract features are not playing a vital role in this case.

With less and limited analytic features, PGN4 demonstrates robustness and efficiency on problem gambling detection. Selecting 5 from 102 features on Dataset A, PGN4 only experiences a 1.7% drop on accuracy and a 0.5% drop on ROC AUC, while Adaboosting drops 7.5% on ROC AUC. Similarly, on Dataset B, applying PGN4 on problem gambling detection with 5 over 27 behavioral features leads to a mere performance drop.

Compared PGN4, the other methods although either have a lower overall performance or have a larger performance dropping from full to limited features, they all confirm the feasibility of predicting problem gambling with few features according to the results in Table 4.

Based on the performance of PGN4, we summarize the top 5 features that lead a compatible detection with full features, as shown in Table 5. Particularly, we discover that live action plays an irreplaceable position in problem gambling detection.

## 4. Discussion

With limited features available, PGN4 is dominant the problem gambling detection compared to other machine learning approaches. This is attributed to the fact that the abstract features distilled by PGN4 from the low-level analytic features significantly enrich the feature space. However, model variation, which results in a tiny scale variation of model performance in every training, is a drawback of PGN4 and all neural network models. The reason is that the randomly initialized neuron weights may lead to a various global minimum during training. Two possible solutions may address this drawback. (1) Increasing the data volume; (2) Increasing the size of minibatch in training.

Table 5 Top 5 features of Dataset A and B

| Dataset | Feature name | Feature discription | Correlation coefficent |
|---|---|---|---|
| Dataset A | NumberofGames31days | Number of games during the first 31 days since the first deposit date | 0.2994 |
| | totalactivedays_31days | Total active days in 31 days since the first deposit date | 0.2916 |
| | p2totalactivedays_31days | Total active days in 31 days since the first deposit date for live action | 0.2835 |
| | playedLA | Played live action odds at least 3 times | 0.2578 |
| | p2SDBets31days | Variability of number of bets per day in live action in 31 days since the first deposit date | 0.2389 |
| Dataset B | bettingdays_liveaction_sqrt | Sum of active betting days: live action: square root transformed | 0.4792 |
| | duration_liveaction_sqrt | Duration of betting days: live action: square root transformed | 0.4714 |
| | bets_per_day_liveaction_sqrt | Bets per betting day: live action: square root transformed | 0.4191 |
| | sum_bets_liveaction_sqrt | Sum of bets: live action: square root transformed | 0.4133 |
| | euros_per_bet_liveaction_sqrt | Euros per bet: live action: square root transformed | 0.3724 |

5. **Conclusion**

In this work, we propose to use 1-D deep neural networks on problem gambling detection to boost the performance with full and limited features. We present a four-convolutional-layer network *PGN4* which is designed based on the available feature size and date volume. Tested on two datasets, PGN4 demonstrates a performance boosting in limited feature space. With only 5 features, PGN4 has the best performance and sustains the detection accuracy and ROC AUC compared with when full features available. Besides, we draw another conclusion that the common top 5 features of two datasets focus on *overall active days*, *overall number of games*, and *live action activities*.

**Reference**

Acion, L., Kelmansky, D., van der Laan, M., Sahker, E., Jones, D., & Arndt, S. (2017). Use of a machine learning framework to predict substance use disorder treatment success. *PloS one*, 12(4), e0175383.

Baker, T. E., Zeighami, Y., Dagher, A., & Holroyd, C. B. (2020). Smoking decisions: Altered reinforcement learning signals induced by nicotine state. *Nicotine and Tobacco Research*, 22(2), 164-171.

Braverman, J., & Shaffer, H. J. (2012). How do gamblers start gambling: Identifying behavioural markers for high-risk internet gambling. *The European Journal of Public Health*, 22(2), 273-278.

Braverman, J., LaPlante, D. A., Nelson, S. E., & Shaffer, H. J. (2013). Using cross-game behavioral markers for early identification of high-risk internet gamblers. *Psychology of Addictive Behaviors*, 27(3), 868.

Deng, X., Lesch, T., & Clark, L. (2019). Applying data science to behavioral analysis of online gambling. *Current Addiction Reports*, 6(3), 159-164.

Gray, H. M., LaPlante, D. A., & Shaffer, H. J. (2012). Behavioral characteristics of Internet gamblers who trigger corporate responsible gambling interventions. *Psychology of Addictive Behaviors*, 26(3), 527.

Gray, H. M., Tom, M. A., LaPlante, D. A., & Shaffer, H. J. (2015). Using opinions and knowledge to identify natural groups of gambling employees. *Journal of gambling studies*, 31(4), 1753-1766.

Griffiths, M. D. (2012). 13 Internet gambling, player protection, and social responsibility. *Routledge international handbook of Internet gambling*, 227.

Jiao, Y., Schneider, B. S. P., Regentova, E., & Yang, M. (2018). Automated quantification of white blood cells in light microscopy muscle images: segmentation augmented by CNN. *In Proceedings of the 2nd International Conference on Vision, Image and Signal Processing,* 1-7.

Jiao, Y., Schneider, B. S. P., Regentova, E., & Yang, M. (2019). DeepQuantify: deep learning and quantification system of white blood cells in light microscopy images of injured skeletal muscles. *Journal of Medical Imaging*, 6(2), 024006-024006.

Kingma, D. P., & Ba, J. (2014). Adam: A method for stochastic optimization. *arXiv preprint arXiv:1412.6980*.

Kiranyaz, S., Avci, O., Abdeljaber, O., Ince, T., Gabbouj, M., & Inman, D. J. (2019). 1D convolutional neural networks and applications: A survey. *arXiv preprint arXiv:1905.03554*.

LeCun, Y., Bengio, Y., & Hinton, G. (2015). Deep learning. *Nature*, 521(7553), 436-444.

Mak, K. K., Lee, K., & Park, C. (2019). Applications of machine learning in addiction studies: A systematic review. *Psychiatry research*.

Rho, M. J., Jeong, J. E., Chun, J. W., Cho, H., Jung, D. J., Choi, I. Y., & Kim, D. J. (2016). Predictors and patterns of problematic Internet game use using a decision tree model. *Journal of behavioral addictions*, 5(3), 500-509.

Rish, I., Bashivan, P., Cecchi, G. A., & Goldstein, R. Z. (2016, March). Evaluating effects of methylphenidate on brain activity in cocaine addiction: a machine-learning approach. In *Medical Imaging 2016: Biomedical Applications in Molecular, Structural, and Functional Imaging* (Vol. 9788, p. 97880O). International Society for Optics and Photonics.

Soussia, M., & Rekik, I. (2018). Unsupervised manifold learning using high-order morphological brain networks derived from T1-w MRI for autism diagnosis. *Frontiers in neuroinformatics*, 12, 70.